\documentclass[letterpaper]{article} % DO NOT CHANGE THIS
\usepackage{aaai2026}
\usepackage{times}  % DO NOT CHANGE THIS
\usepackage{helvet}  % DO NOT CHANGE THIS
\usepackage{courier}  % DO NOT CHANGE THIS
\usepackage[hyphens]{url}  % DO NOT CHANGE THIS
\usepackage{graphicx} % DO NOT CHANGE THIS
\usepackage{amsfonts} % For \mathbb
\usepackage{amsmath}
\usepackage{multirow}  
\usepackage{booktabs}  
\usepackage[table]{xcolor} 
\usepackage{amssymb}
\usepackage{natbib}  % DO NOT CHANGE THIS AND DO NOT ADD ANY OPTIONS TO IT
\usepackage{caption} % DO NOT CHANGE THIS AND DO NOT ADD ANY OPTIONS TO IT
\usepackage{algorithm}
\usepackage{algorithmic}

\usepackage{newfloat}
\usepackage{listings}

\newcommand{\ie}{\textit{i}.\textit{e}.}
\newcommand{\eg}{\textit{e}.\textit{g}.}

\DeclareCaptionStyle{ruled}{labelfont=normalfont,labelsep=colon,strut=off} % DO NOT CHANGE THIS
\floatstyle{ruled}
\newfloat{listing}{tb}{lst}{}
\floatname{listing}{Listing}
\title{Learning to Tell Apart:  Weakly Supervised Video Anomaly Detection via Disentangled Semantic Alignment}
\author {
    Wenti Yin\textsuperscript{\rm 1},
    Huaxin Zhang\textsuperscript{\rm 1},
    Xiang Wang\textsuperscript{\rm 1},
    Yuqing Lu\textsuperscript{\rm 1},
    Yicheng Zhang\textsuperscript{\rm 1},
    Bingquan Gong\textsuperscript{\rm 1},\\
    Jialong Zuo\textsuperscript{\rm 1},
    Li Yu\textsuperscript{\rm 2},
    Changxin Gao\textsuperscript{\rm 1},
    Nong Sang\textsuperscript{\rm 1}\thanks{Corresponding author.}
}
\affiliations {
    \textsuperscript{\rm 1}Key Laboratory of Image Processing and Intelligent Control, School of Artificial Intelligence and Automation,\\
    Huazhong University of Science and Technology\\
    \textsuperscript{\rm 2}School of Electronic Information and Communications, Huazhong University of Science and Technology\\
    \{yinwt, nsang\}@hust.edu.cn
}
\begin{document}

\maketitle
% ----------- Abstract -----------
\begin{abstract}
Recent advancements in weakly-supervised video anomaly detection have achieved remarkable performance by applying the multiple instance learning paradigm based on multimodal foundation models such as CLIP to highlight anomalous instances and classify categories.
However, their objectives may tend to detect the most salient response segments, while neglecting to mine diverse normal patterns separated from anomalies, and are prone to category confusion due to similar appearance, leading to unsatisfactory fine-grained classification results.
Therefore, we propose a novel Disentangled Semantic Alignment Network (DSANet) to explicitly separate abnormal and normal features from coarse-grained and fine-grained aspects, enhancing the distinguishability.
Specifically, at the coarse-grained level, we introduce a self-guided normality modeling branch that reconstructs input video features under the guidance of learned normal prototypes, encouraging the model to exploit normality cues inherent in the video, thereby improving the temporal separation of normal patterns and anomalous events.
At the fine-grained level, we present a decoupled contrastive semantic alignment mechanism, which first temporally decomposes each video into event-centric and background-centric components using frame-level anomaly scores and then applies visual-language contrastive learning to enhance class-discriminative representations.
Comprehensive experiments on two standard benchmarks, namely XD-Violence and UCF-Crime, demonstrate that DSANet outperforms existing state-of-the-art methods. 
% Code is available at \url{https://github.com/lessiYin/DSANet}.
\end{abstract}

\begin{links}
    \link{Code}{https://github.com/lessiYin/DSANet}
    % \link{Datasets}{https://aaai.org/example/datasets}
    % \link{Extended version}{https://aaai.org/example/extended-version}
\end{links}

% ----------- Introduction -----------
\section{Introduction}
\label{sec:introduction}

\begin{figure}[t]
    
    \centering
    
    \includegraphics[width=1.0\columnwidth]{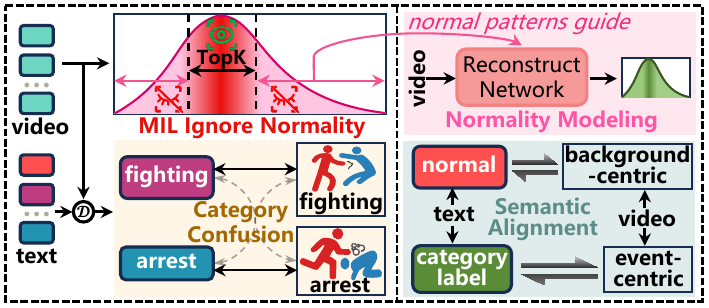}
    
    \caption{
    Schematic diagram about motivation.
    We identify two main issues: \emph{1) limited understanding of normality},
    and \emph{2) category confusion}. We address them through normality modeling and decoupled contrastive semantic alignment.
    % Motivation and Proposed Solutions Overview. 
    % We identify two main issues—limited understanding of normality and category confusion—and address them via normality modeling and decoupled contrastive semantic alignment.
    }
    \label{fig:motivation2}
\end{figure}

%%% Context
Weakly Supervised Video Anomaly Detection (WS-VAD)~\cite{sultani2018real} aims to temporally detect anomaly segments in a long untrimmed video with only video-level labels (\ie, indicating whether a video contains an anomaly), drastically reducing annotation costs compared to its fully supervised counterparts~\cite{wu2024deep,abdalla2024video,nayak2021comprehensive}, and has received considerable attention in recent years~\cite{Wang_2021_ICCV, Wang_2022_CVPR, shi2025shield, wang2025sdeval, zhu2024video, liang2023video}.
The predominant approach in WS-VAD is built upon the multiple instance learning (MIL) framework~\cite{tian2021weakly,lv2023unbiased,chen2024prompt}. 
The general pipeline involves first extracting deep features for each video using a pre-trained backbone like I3D~\cite{carreira2017quo} and CLIP~\cite{radford2021learning}, and then feeding the obtained features to a binary classifier to generate instance-level anomaly scores~\cite{Yu_2025_CVPR, wang2025affordance, wang2025dag}.
For example,  CLIP-TSA~\cite{joo2023clip} uses CLIP’s visual encoder with multi-scale temporal aggregation and a multiple instance learning branch for detection. VadCLIP~\cite{wu2024vadclip} employs a binary classifier for anomaly detection and text prompts for anomaly types identification. PEMIL~\cite{pu2024learning} designs anomaly- and context-aware prompts to model complex event boundaries. ITC~\cite{liu2024injecting} introduces learnable textual cues in a dual-branch framework for robust cross-modal anomaly recognition.

%%% Challenge
Despite its recent success, the prevailing WS-VAD approaches based on multiple instance learning still suffer from two fundamental limitations. At the coarse-grained level, the discriminative nature of MIL results in an incomplete understanding of normality. By focusing exclusively on identifying the most salient anomalous segments, such models fail to construct a robust and explicit representation of the diverse normal patterns present within a video. This deficiency compromises the model’s ability to distinguish between true anomalies and complex yet benign events, leading to ambiguous decision boundaries and an increased false positive rate.
At the fine-grained level, anomaly classification remains challenging due to the possible visual similarity among different abnormal events and between anomalies and normal contexts.
In the absence of frame-level supervision, models often confuse co-occurring background patterns with true anomalies.
The learned representations for different anomaly categories often become entangled in the embedding space, leading to semantic confusion and reduced inter-class separability. This confusion ultimately limits the model’s ability to perform accurate anomaly categorization and discriminative localization.

%%% Contribution
To address these challenges, we propose DSANet that enhances WS-VAD through synergistic normality modeling and disentangled contrastive semantic alignment, as shown in Figure~\ref{fig:motivation2}.
% % SG-NM
To resolve the lack of explicit normality understanding, we introduce a Self-Guided Normality Modeling (SG-NM) branch. Inspired by the observation that even within anomalous sequences, local regions still exhibit intrinsic normality~\cite{luo2025exploring} (\eg~normal backgrounds with visual consistency), SG-NM dynamically mines a compact set of normal prototypes directly from each input video, without relying on an external memory bank~\cite{zhou2023dual}.
These prototypes supervise a reconstruction objective that drives the model to learn video-specific normal characteristics in a generative manner.
This design enhances the discriminative detector with an internal understanding of normality, 
enabling the model to better distinguish normal behavior and more accurately identify anomalies.
The SG-NM branch is self-contained, memory-free, and requires no dataset-level priors, making it scalable and data-efficient.
% DCSA
To mitigate category confusion caused by visual similarities between different anomaly classes and normal contexts, we design a Decoupled Contrastive Semantic Alignment (DCSA) mechanism. Guided by initial anomaly predictions, DCSA explicitly disentangles video features into an event-centric and a background-centric prototype. A dual contrastive objective then aligns the event prototype with its corresponding anomaly class, while consistently aligning the background prototype with a universal ``normal" class. This disentanglement yields more discriminative representations, enhancing the model's ability for fine-grained anomaly classification and precise temporal localization.
%%% Claim

%%% Conclusion
In summary, our main contributions are as follows:
\begin{itemize}
\item We introduce \textbf{Self-Guided Normality Modeling}, a generative reconstruction module that dynamically mines video-specific normal patterns without external memory. By incorporating normality cues overlooked by MIL-based paradigms, it facilitates more accurate and complete coarse-grained anomaly detection.
\item We propose \textbf{Decoupled Contrastive Semantic Alignment}, a mechanism that explicitly separates event- and background-centric prototypes and aligns them with respective semantic targets, addressing semantic confusion and improving fine-grained anomaly classification.
\item Extensive experiments verify the effectiveness of the proposed \textbf{DSANet}, which outperforms existing state-of-the-art methods on two standard WS-VAD benchmarks.
\end{itemize}

% ----------- Related Work -----------
\section{Related Work}

\subsection{Vision-Language Pre-training}
Vision-Language Pre-training (VLP) has become a dominant paradigm for learning joint representations from large-scale image-text data, enabling remarkable zero-shot transfer capabilities~\cite{ho2025review}. Pioneering works such as CLIP~\cite{radford2021learning} and ALIGN~\cite{jia2021scaling} use dual-encoder architectures trained with contrastive loss to align visual and textual embeddings.
Recent trends aim to improve capability and efficiency.
One line adopts unified encoder-decoder frameworks that jointly handle understanding and generation tasks, exemplified by BLIP~\cite{li2022blip}, which introduced a data bootstrapping method to denoise web captions, and CoCa~\cite{yu2022cocacontrastivecaptionersimagetext}, which combines contrastive and captioning losses. Another focuses on parameter efficiency by leveraging powerful, frozen uni-modal models. Flamingo~\cite{alayrac2022flamingo} bridges frozen vision encoders and large language models with a Perceiver Resampler and trainable gated cross-attention layers, while BLIP-2~\cite{li2023blip} introduced a lightweight Querying Transformer to link frozen components with minimal training cost. VLP models have been widely applied to downstream tasks such as text-video retrieval~\cite{wang2024text,tian2024holistic}, visual question answering~\cite{zou2024language,li2024configure}, and open-vocabulary action recognition~\cite{huang2024frosterfrozenclipstrong,jia2023generatingactionconditionedpromptsopenvocabulary}. In this work, following previous practices~\cite{wu2024vadclip,dev2024reflip}, we construct DSANet based on CLIP~\cite{radford2021learning} for weakly supervised video anomaly detection.

\subsection{Weakly Supervised Video Anomaly Detection}
This task was first formalized by Sultani et al.~\cite{sultani2018real} as a multiple instance learning (MIL) problem, introducing a deep ranking framework that enables the model to assign higher anomaly scores to abnormal segments under weak supervision. Subsequent works have expanded and refined this paradigm~\cite{tian2021weakly,zanella2024harnessing,zhang2025holmes}. To enhance temporal modeling, RTFM~\cite{tian2021weakly} integrates temporal convolution and self-attention to capture multi-scale temporal dependencies. To mitigate contextual bias in MIL, UMIL~\cite{lv2023unbiased} introduces a strategy to learn stable representations across ``confident" and ``ambiguous" samples, thereby improving classifier robustness. To address the weakness of the supervision signal, Feng et al.~\cite{feng2021mist} propose a two-stage self-training framework that refines discriminative feature representations by using a multiple instance pseudo-label generator to train a self-guided attention encoder.
With the rise of vision-language pre-training (VLP) models, WS-VAD has been transitioning from traditional statistical pattern recognition towards semantic-aligned reasoning. 
Early approaches adopted VLP models (\eg, CLIP) as visual feature extractors~\cite{joo2023clip}. Recent efforts~\cite{wu2024vadclip,pu2024learning,liu2024injecting} incorporate prompt-based or learnable textual cues into the MIL framework to facilitate anomaly type recognition and enhance event-level discrimination.

% ----------- Method -----------

\begin{figure*}[t]
    \centering
    
    \includegraphics[width=1.0\textwidth]{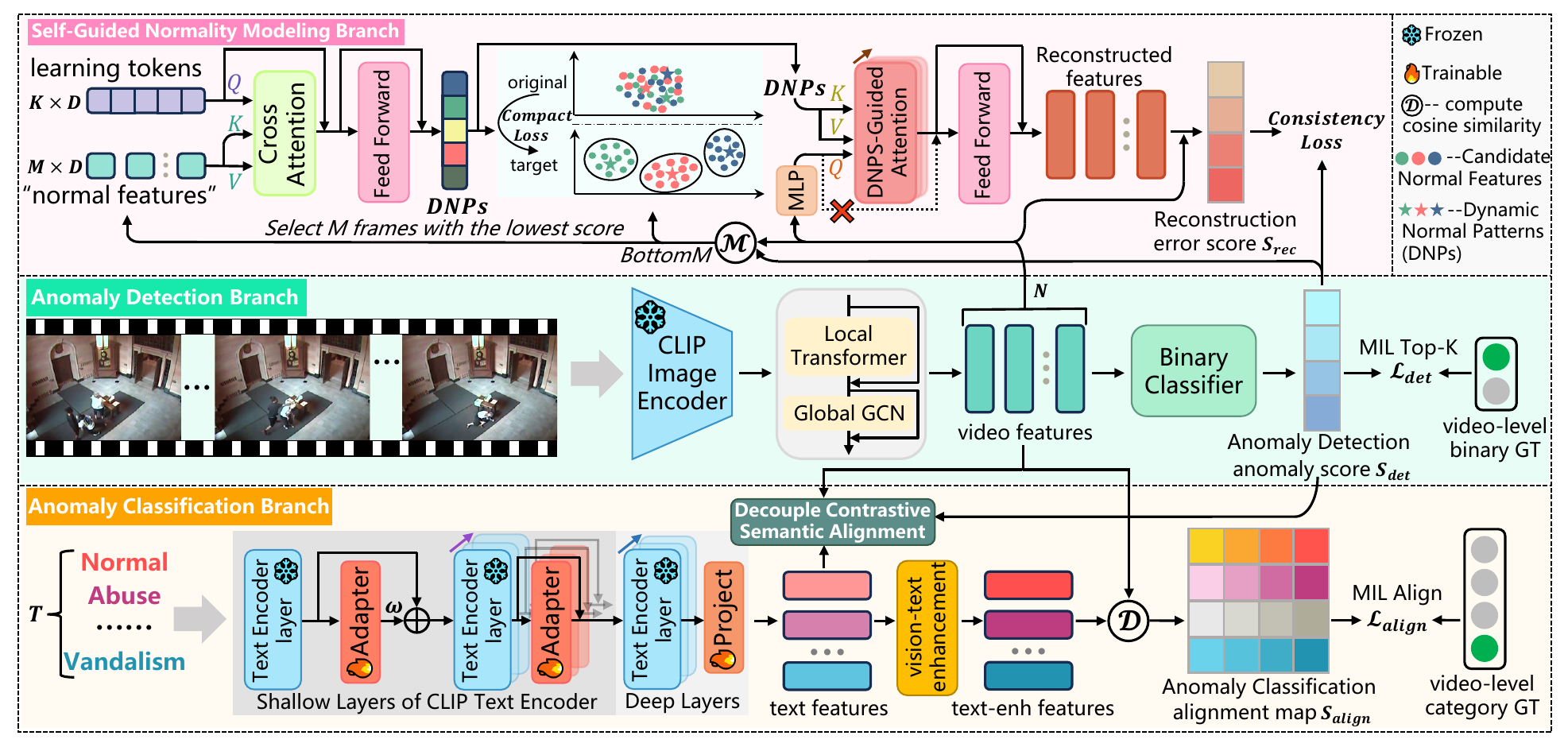}
    \caption{
    Overview of the proposed DSANet. The model consists of three collaborative branches. The Anomaly Detection Branch produces initial  frame-level binary scores using a MIL framework. The Self-Guided Normality Modeling Branch enhances the model's understanding of normal patterns by mining Dynamic Normal Patterns within the video to guide feature reconstruction, improving its ability to distinguish normal from abnormal. The Anomaly Classification Branch aligns video features with textual category embeddings for fine-grained classification, using Lightweight Text Adapters for adaptation and a Decoupled Contrastive Semantic Alignment mechanism to distinguish various anomaly types from normal categories.
    }
    \label{fig:framework}
\end{figure*}
\section{Methodology}

\subsection{Overview}
\subsubsection{Problem Definition.}
We address the task of WS-VAD, where the input is a video set $\mathcal{V} = \{v_i\}_{i=1}^n$. Each video $v_i$ has video-level labels: (i) a binary anomaly indicator $y_i \in \{0,1\}$ and (ii) an anomaly category label $ c_i \in \{0,1,\dots,C-1\}$. A video is normal if $y_i = 0$ and $c_i = 0$; otherwise, it contains at least one anomaly and $c_i$ denotes its category. 
Under weak supervision, two subtasks exist: coarse-grained WS-VAD assigns anomaly scores to frames, while fine-grained WS-VAD identifies categories and localizes anomalies.

\subsubsection{Overall Framework.}
As illustrated in Figure~\ref{fig:framework}, our model has three branches: Anomaly Detection, Self-Guided Normality Modeling (SG-NM), and Anomaly Classification.

\textbf{Anomaly Detection Branch} generates coarse-grained, binary frame-level anomaly scores based on the multiple instance learning (MIL) framework. Following~\cite{wu2024vadclip}, we use a frozen CLIP image encoder to extract frame-wise features $F_{clip} \in \mathbb{R}^{N \times D}$, where $N$ is the number of frames and $D$ is the feature dimension. To augment the CLIP features with temporal information, the frame-wise features are fed into a two-stage temporal modeling module: (1) a local Transformer, which restricts attention to non-overlapping temporal windows to capture short-range dynamics, and (2) a dual-channel GCN for long-range dependencies, whose edges are constructed based on feature cosine similarity and relative temporal distance.
The temporally contextualized video representation $F_{{video}} \in \mathbb{R}^{N \times D}$ is then passed through a binary classifier to produce coarse-grained anomaly scores $S_{det} \in \mathbb{R}^{N \times 1}$.
Finally, frame-level anomaly probabilities $p_i = \sigma(s_i)$ are aggregated by averaging top-$k$ scores for video-level prediction $\bar{p}$, with loss:
\begingroup
\footnotesize
\begin{equation}
\mathcal{L}_{det} = - \sum_{c \in \{0,1\}} y_c \log \bar{p}^c (1 - \bar{p})^{1-c}, \quad 
\bar{p} = \frac{1}{k} \sum_{i \in \mathcal{I}_{\text{top-}k}} p_i,
\end{equation}
\endgroup
where $\mathcal{I}_{\text{top-}k}$ denotes the indices of the top-$k$ scoring frames.

\textbf{SG-NM Branch} addresses the deficiency in normality modeling within the Anomaly Detection Branch, whose discriminative MIL objective tends to focus on the most salient anomalous segments while neglecting the diverse and informative normal patterns in the video.
Specifically, we mine the video-specific normal prototypes to guide a reconstruction learning process. It computes a reconstruction-based anomaly score $S_{rec}$, which is aligned with the anomaly detection score $S_{det}$, forming a self-distillation mechanism that strengthens the model’s normal representation.

\textbf{Anomaly Classification Branch} enables fine-grained category-aware anomaly detection through visual-text alignment. 
In existing works~\cite{wu2024vadclip}, all category labels (\eg, \textit{`Abuse'}) are embedded into the text feature space $T_{text} \in \mathbb{R}^{C \times D}$ using the frozen CLIP text encoder. A frame-category alignment map $S_{align} \in \mathbb{R}^{N \times C}$ is then computed between the category embeddings and video features, and supervised by the category label under the MIL strategy:
the top-$k$ values in each column $M_{:,c}$ are averaged to get video-level category scores $S_c$. The alignment loss $\mathcal{L}_{align}$ is defined as the multi-class cross-entropy between predicted $p$ from $S_c$ and the one-hot category label vector:
\begingroup
\footnotesize
\begin{equation}
\mathcal{L}_{align} = - \sum_{c = 0}^{C-1} y_c \log \left( \bar{p}_c \right),\quad \bar{p}_c = \text{softmax}\left(S_c/ \tau\right),
\label{eq:Lalign}
\end{equation}
\endgroup
where $S_c = \frac{1}{k} \sum_{i \in \mathcal{I}_{\text{top-}k}^{(c)}} M_{i,c}$. Following~\cite{wu2024vadclip}, before computing the alignment map, we use a Vision-Text Enhance module that injects video-derived visual cues into the text embedding to enhance event-specific semantics.
Technically, frame-level features $F_{video}$ are first aggregated into a global video representation $V$ using anomaly score-guided weighting: $V=\mathrm{Norm}(S_{det}^\top F_{video})$, where $\mathrm{Norm}(\cdot)$ denotes L2 normalization. This visual cue $V$ is then fused with $T_{text}$ via a skip-connected feed-forward network to generate enhanced category embeddings $T_{text-enh}$.

To reduce semantic confusion between visually similar anomalies and normal contexts, we introduce a 
\textbf{Decoupled Contrastive Semantic Alignment} mechanism to decouple visual features for category-aware alignment. We also propose \textbf{Lightweight Text Adapters} to obtain domain-adaptive text representations. We next detail the proposed modules.
% We next provide detailed descriptions of the proposed modules.
% To adapt textual features beyond coarse labels, we introduce \textbf{Lightweight Text Adapters} for efficient and domain-adaptive representation learning.
% \wx{removed by wx}
% To reduce semantic confusion frome visually similar anomalies and adjacent normal contexts, we propose a \textbf{Decoupled Contrastive Semantic Alignment} mechanism: guided by $S_{det}$, visual features are decoupled into event-centric and background-centric prototypes, which are respectively aligned with the corresponding anomaly category feature and the normal category feature. Furthermore, we found that coarse-grained category text labels lack detailed anomaly semantic information, which increases the difficulty of aligning with visual features, and this phenomenon has also been observed in~\cite{pu2024learning}. To obtain domain-adaptive text representations, we propose \textbf{Lightweight Text Adapters} to enable the update of text representations in an adaptive and computationally efficient manner.

\subsection{Self-Guided Normality Modeling}
\label{sec:SGNM}
The MIL-based Anomaly Detection Branch focuses on only the most salient video parts (\ie, top-$k$ frames), leading to an incomplete understanding of rich normal patterns and blurring the boundaries between normality and anomaly.
In contrast, the SG-NM branch, which operates only during training, explicitly models video-specific normality to guide video feature reconstruction. 
This reconstruction pathway complements the detection branch by leveraging inherent normal cues in the video, improving the temporal separation between normal and anomalous events.

We first extract representative normal patterns directly from the input video to ensure contextual relevance and adaptability, rather than relying on external memory banks to model dataset-level normality~\cite{zhou2023dual}.
Specifically, we use initial anomaly scores $S_{det}$ from the detection branch to select the frame features with the bottom-$M$ lowest scores, forming the candidate normal feature set $F_{n} \in \mathbb{R}^{M \times D}$.
To extract representative normal patterns from $F_{n}$, we apply a single-layer cross-attention module. A set of $K=16$ learnable queries $Q_{learn} \in \mathbb{R}^{K \times D}$ attends to $F_{n}$ (used as both key and value), extracting $K$ distilled Dynamic Normal Patterns (DNPs), denoted as $P \in\mathbb{R}^{K \times D}$.
To ensure that DNPs purely represent normal features and resist contamination from anomalies within the candidate set, we introduce a Normalcy Concentration Loss $\mathcal{L}_{compact}$. 
This loss encourages each feature in $F_{n}$ to be close to at least one DNP.
It is a self-supervised objective that forces the DNPs to become compact and representative centers of normal patterns.
The loss is defined as the average minimum distance from each feature in $F_{n}$ to the set of DNPs:
% Normalcy Concentration Loss
\begingroup
\footnotesize
\begin{equation}
\mathcal{L}_{compact} = \frac{1}{M} \sum_{i=1}^M \min_{j \in \{1, \dots, K\}} d\big(F_{n}(i), P(j)\big),
\end{equation}
\endgroup
where \( d(\cdot, \cdot) \) represents the cosine distance, \( F_{n}(i) \) is the \( i \)-th feature in the candidate set, and \( P(j) \) is the \( j \)-th DNP.

Guided by the extracted normal patterns, we then design a multi-layer cross-attention decoder that reconstructs the video feature for anomaly detection.
Video features $F_{video} \in \mathbb{R}^{N \times D}$ are first mapped via an MLP and used as the query, while DNPs $P$ serve as key and value. To prevent anomaly leakage, the first attention layer excludes residual connections, ensuring that reconstruction relies solely on normal patterns.
Each decoder layer applies cross-attention between transformed video features and the DNPs, followed by a feed-forward network. The final reconstructed features $F_{rec}$ are compared to the original features $F_{video}$, and a frame-level anomaly score $S_{rec} \in \mathbb{R}^{N \times 1}$ is obtained by computing their cosine distance, normalized to $[0,1]$. A higher reconstruction error indicates greater deviation from normalcy.

To unify the discriminative and generative anomaly views ($S_{det}$ and $S_{rec}$), we introduce a consistency loss $\mathcal{L}_{consist}$ based on mean squared error. This acts as a form of knowledge distillation, where each branch guides the other.
\begingroup
\footnotesize
\begin{equation}
\mathcal{L}_{consist} = \frac{1}{N} \sum_{i=1}^{N} \left( S_{det}(i) - S_{rec}(i) \right)^2.
\end{equation}
\endgroup
This consistency objective encourages mutual refinement between branches by combining MIL's focus on anomalous segments with the SG-NM's modeling of intrinsic normality, promoting a holistic understanding of video anomalies.

\subsection{Decoupled Contrastive Semantic Alignment}

\begin{figure}[tb]
    
    \centering
    
    \includegraphics[width=1.0\columnwidth]{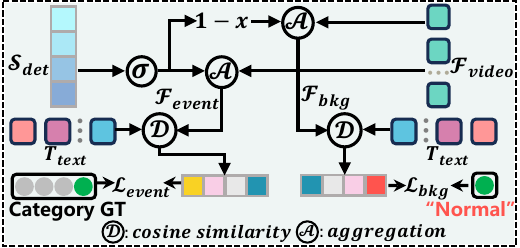}
    
    \caption{
    Detailed structure of the proposed Decoupled Contrastive Semantic Alignment module.
    }
    \label{fig:DCSA}
\end{figure}

Fine-grained anomaly detection under weak supervision is challenged by visual similarity among anomaly types and interference from surrounding normal frames, leading to semantic confusion and poor class separability.
% as the model may associate irrelevant contextual patterns with anomaly semantics.
To resolve this, we propose a Decoupled Contrastive Semantic Alignment (DCSA) mechanism shown in Figure~\ref{fig:DCSA} for fine-grained cross-modal alignment between event-background decoupled visual features and text features, thus enhancing category discrimination and temporally localization accuracy.

Specifically, we first decouple visual features.
Given video features $F_{video}$, we compute event-centric prototype $F_{event}$ and background-centric prototype $F_{bkg}$ using frame-level anomaly scores $S_{det}$ produced by the detection branch:
\begingroup
\footnotesize
\begin{equation}
F_{event} = w_{event}^\top F_{video}, \quad F_{bkg} = w_{bkg}^\top F_{video},
\label{eq:dcsa}
\end{equation}
\endgroup
where $w_{event} = \text{Softmax}(S_{det}), w_{bkg} = 1 - w_{event}$.
% This generates two disentangled features $F_{event}, F_{bkg} \in \mathbb{R}^{D}$, corresponding to salient events and contextual background.
This performs a context-preserving soft decoupling via weighted aggregation of all frame features, shifting the focus to either salient events ($F_{event}$) or the background ($F_{bkg}$).

Subsequently, we decouple text features.
The class text embeddings \( T_{text} = \{ t_0, t_1, \dots, t_{C-1} \} \) represent the ``normal'' class $( t_0 )$ and the $C-1$ anomaly classes $( t_1, \dots, t_{C-1} )$. To promote a clear margin in the semantic space, we use a separation loss that pushes the ``normal" class embedding $t_0$ away from all abnormal class embeddings $\{t_a \mid a \in \{1, ..., C-1\}\}$. This is achieved by minimizing the absolute cosine similarities:
\begingroup
\footnotesize
\begin{equation}
\mathcal{L}_{sep} = \sum_{a=1}^{C-1} \left| \frac{t_0^\top t_a}{\|t_0\| \|t_a\|} \right|.
\end{equation}
\endgroup
% This encourages semantic decoupling between normal and abnormal categories.

Finally, we apply fine-grained cross-modal semantic alignment to enhance class-discrimination.
Given a video label \( y \in \{ 0, 1, \dots, C-1 \} \), we define a dual contrastive alignment loss as:
\begingroup
\footnotesize
\begin{equation}
\mathcal{L}_{dcsa} = \mathcal{L}_{event} + \mathcal{L}_{bkg}.
\end{equation}
\endgroup
The event alignment loss encourages \( F_{event} \) to align with its ground-truth class \( t_c \), while contrasting it against others:
\begingroup
\footnotesize
\begin{equation}
\mathcal{L}_{event} = 
- \sum_{c=0}^{C-1} {y}_c \log \frac{
\exp \left( \text{sim}(F_{event}, t_c) / \tau \right)
}{
\sum\limits_{j=0}^{C-1} \exp \left( \text{sim}(F_{event}, t_j) / \tau \right)
},
\label{eq:Levent}
\end{equation}
\endgroup
% \begingroup
% \footnotesize
% \begin{equation}
% \mathcal{L}_{event} = 
% - \log \frac{
% \exp \left( \text{sim}(F_{event}, t_y) / \tau \right)
% }{
% \sum\limits_{c=0}^{C-1} \exp \left( \text{sim}(F_{event}, t_c) / \tau \right)
% }.
% \label{eq:Levent}
% \end{equation}
% \endgroup
The background alignment loss consistently aligns \( F_{bkg} \) with the ``normal" embedding \( t_0 \), serving as a semantic regularizer to encourage a stable representation for normality:
\begingroup
\footnotesize
\begin{equation}
\mathcal{L}_{bkg} = 
- \log \frac{
\exp \left( \text{sim}(F_{bkg}, t_0) / \tau \right)
}{
\sum\limits_{c=0}^{C-1} \exp \left( \text{sim}(F_{bkg}, t_c) / \tau \right)
}.
\label{eq:Lbkg}
\end{equation}
\endgroup
Here, \( \text{sim}(\cdot, \cdot) \) denotes cosine similarity. This decoupled formulation addresses both abnormal and normal cases:
\begin{itemize}
  \item \text{Abnormal video} (\( y = k \neq 0 \)):  
  \( F_{\text{event}} \leftrightarrow t_k \), \( F_{\text{bkg}} \leftrightarrow t_0. \)

  \item \text{Normal video} (\( y = 0 \)):  
  \( F_{\text{event}} \leftrightarrow t_0 \), \( F_{\text{bkg}} \leftrightarrow t_0. \)
\end{itemize}

This decoupled contrastive semantic alignment helps disentangle anomaly-relevant features from shared contextual patterns, which reduces cross-category semantic confusion and enhances class-discriminative representations.

\subsection{Lightweight Text Adapter}
To enable domain-specific adaptation while preserving the general knowledge of CLIP, we insert lightweight adapters into the early Transformer blocks of the text encoder. Specifically, adapters are placed in the first $L$ layers, and a fusion weight $\omega_t$ controls the contribution of the adapted features. Each adapter operates in parallel with the original self-attention and feed-forward layers. Given the intermediate feature $x$, the adapter produces an output $x_{adapt}$, which is fused with the original as follows:
\begingroup
\footnotesize
\begin{equation}
x_{out} = (1 - \omega_t) \cdot x + \omega_t \cdot \text{Norm}(x_{adapt}).
\end{equation}
\endgroup
This modulation preserves the original feature norm while allowing directional adaptation in the embedding space.

\subsection{Training and Inference.}
\subsubsection{Training.}
The model is optimized with a unified loss balancing learning objectives across all branches:
\begingroup
\footnotesize
\begin{equation}
\begin{aligned}
\mathcal{L}_{total} =\ & \mathcal{L}_{det} + \lambda \mathcal{L}_{align} + \mathcal{L}_{consist} + \\
&\mathcal{L}_{compact} + \mathcal{L}_{dcsa} + \mathcal{L}_{sep},
\end{aligned}
\end{equation}
\endgroup
where $\lambda$ balances different loss terms.
\subsubsection{Inference.}
For coarse-grained WS-VAD, we directly use $S_{det}$ from the Anomaly Detection Branch, which provides a reliable frame-level binary confidence for detecting anomalies.
For fine-grained prediction, we propose a Hierarchical Belief Modulation strategy that integrates temporal cues from the Anomaly Detection Branch's $S_{det}$ with semantic cues from the Anomaly Classification Branch's $S_{align}$.
Specifically, $S_{det}$ serves as a temporal prior. This anomaly score is then distributed over the $C$ fine-grained classes according to their relative likelihoods predicted by $S_{align}$.
This design ensures that the final predictions are anchored to the robust temporal boundaries identified by the Anomaly Detection Branch, while the Anomaly Classification Branch's expertise is precisely targeted at resolving ambiguity between different anomaly classes. 
The process is calibrated by a single hyperparameter, the temperature ratio $\beta$.

% ----------- Experiments -----------
\section{Experiments}

\subsection{Experimental Setup}

\noindent\textbf{Datasets and Evaluation Metrics.}
We evaluate our method on two widely-used WS-VAD benchmarks: XD-Violence~\cite{wu2020not} and UCF-Crime~\cite{sultani2018real}, both providing only video-level labels for training. For coarse-grained evaluation, we follow standard protocols using frame-level AUC for UCF-Crime and Average Precision (AP) for XD-Violence. For fine-grained evaluation, we follow previous work~\cite{wu2024vadclip}, computing mean Average Precision (mAP) under IoU thresholds of 0.1 to 0.5 with step 0.1, and report average mAP (AVG).

\noindent\textbf{Implementation Details.}
To ensure a fair comparison with existing methods~\cite{wu2024vadclip,dev2024reflip}, the frozen CLIP (ViT-B/16) is adopted to extract features.
Temperature scaling factors in Eq. (\ref{eq:Lalign}), (\ref{eq:Levent}) and (\ref{eq:Lbkg}) are set to 0.07. The text adapter is set to $L=3$ with fusion weight $\omega_t=0.1$ for UCF-Crime and $L=1$, $\omega_t=0.6$ for XD-Violence. An 8-layer cross-attention decoder is used in the SG-NM branch, where $M$ is set to 80\% of the video length to select candidate normal frames. 
During inference, the temperature ratio $\beta$ is set to $5.0$ on UCF-Crime and $1.0$ on XD-Violence. The loss weight $\lambda$ for the unified objective is $1.1$ on UCF-Crime and $5.0$ on XD-Violence. We use the AdamW optimizer with batch sizes of 64 (UCF-Crime) and 96 (XD-Violence) to optimize the model. All experiments are conducted on a single NVIDIA 4090 GPU using PyTorch. Training lasts 10 epochs with learning rates of $7\mathrm{e}{-5}$ for UCF-Crime and $1\mathrm{e}{-5}$ for XD-Violence.

\begin{table}[t]
    \centering
    
    \small
    
    \setlength{\tabcolsep}{1.8pt} 
    
    \begin{tabular}{c|lcc}
        \toprule 
        Category & Method & Features & AP(\%) \\
        \midrule 
        \multirow{1}{*}{Un} & LTR\cite{hasan2016learning} & - & 30.77 \\
        \midrule 
        \multirow{12}{*}{Weak} & RAD\cite{sultani2018real} & C3D & 73.20 \\
        & RTFML\cite{tian2021weakly} & I3D & 77.81 \\
        & ST-MSL\cite{li2022self} & I3D & 78.28 \\
        & LA-Net\cite{pu2022locality} & I3D & 80.72 \\
        & DMU\cite{zhou2023dual} & I3D & 82.41 \\
        & PEL4VAD\cite{pu2024learning} & I3D & 85.59 \\
        \cline{2-4}
        & CLIP-TSA\cite{joo2023clip} & CLIP & 82.19 \\
        & TPWNG\cite{yang2024text}  & CLIP & 83.68 \\
        & VadCLIP\cite{wu2024vadclip}  & CLIP & 84.51 \\
        & ITC\cite{liu2024injecting}  & CLIP & 85.45 \\
        & ReFLIP\cite{dev2024reflip}  & CLIP & 85.81 \\
        
        & \textbf{DSANet(Ours)} %\cellcolor{gray!20}
        & \textbf{CLIP} %\cellcolor{gray!20}
        & \textbf{86.95} \\%\cellcolor{gray!20}
        \bottomrule
    \end{tabular}
    % \vspace{-2mm}
    \caption{Coarse-grained comparisons on XD-Violence.}
    \label{tab:cg_XD}
\end{table}

\begin{table}[t]
    \centering
    \small
    \setlength{\tabcolsep}{1.0pt}
    \begin{tabular}{c|lcc}
        \toprule 
        Category & Method & Features & AUC(\%) \\
        \midrule 
        \multirow{1}{*}{Un} & LTR\cite{hasan2016learning} & - & 50.60 \\
        \midrule 
        \multirow{12}{*}{Weak} & RAD\cite{sultani2018real} & I3D & 77.92 \\
        & RTFML\cite{tian2021weakly} & I3D & 84.30 \\
        & LA-Net\cite{pu2022locality} & I3D & 85.12 \\
        & ST-MSL\cite{li2022self} & I3D & 85.30 \\
        & DMU\cite{zhou2023dual} & I3D & 86.75 \\
        & PEL4VAD\cite{pu2024learning} & I3D & 86.76 \\
        \cline{2-4}
        & CLIP-TSA\cite{joo2023clip} & CLIP & 87.58 \\
        & TPWNG\cite{yang2024text}  & CLIP & 87.79 \\
        & VadCLIP\cite{wu2024vadclip}  & CLIP & 88.02 \\
        & ReFLIP\cite{dev2024reflip}  & CLIP & 88.57 \\
        & ITC\cite{liu2024injecting}  & CLIP & 89.04 \\
        
        & \textbf{DSANet(Ours)} %\cellcolor{gray!20}
        & \textbf{CLIP} %\cellcolor{gray!20}
        & \textbf{89.44} \\%\cellcolor{gray!20}
        \bottomrule
    \end{tabular}
    % \vspace{-2mm}
    \caption{Coarse-grained comparisons on UCF-Crime.}
    \label{tab:cg_UCF}
\end{table}

\begin{table}[t]
    \centering
    
    \small
    
    \setlength{\tabcolsep}{4.0pt}

    \begin{tabular}{lcccccc}
        \toprule
        \multirow{2}{*}{Method} & \multicolumn{6}{c}{mAP@IOU(\%)} \\
        \cmidrule(lr){2-7}
        & 0.1 & 0.2 & 0.3 & 0.4 & 0.5 & AVG \\
        \midrule
        RAD(2018) & 22.72 & 15.57 & 9.98 & 6.20 & 3.78 & 11.65 \\
        AVVD(2022) & 30.51 & 25.75 & 20.18 & 14.83 & 9.79 & 20.21 \\
        VadCLIP(2024) & 37.03 & 30.84 & 23.38 & 17.90 & 14.31 & 24.70 \\
        ITC(2024) & 40.83 & 32.80 & 25.42 & 19.65 & 15.47 & 26.83 \\
        ReFLIP(2024) & 39.24 & 33.45 & 27.71 & 20.86 & 17.22 & 27.36 \\
        % \rowcolor{gray!20}
        \textbf{DSANet(Ours)} & \textbf{40.93} & \textbf{34.63} & \textbf{28.21} & \textbf{22.70} & \textbf{17.89} & \textbf{28.87} \\
        \bottomrule
    \end{tabular}
    % \vspace{-2mm}
    \caption{Fine-grained comparisons on XD-Violence.}
    \label{tab:fg_XD}
\end{table}

\begin{table}[t]
    \centering
    
    \small
    
    \setlength{\tabcolsep}{4.7pt}

    \begin{tabular}{lcccccc}
        \toprule
        \multirow{2}{*}{Method} & \multicolumn{6}{c}{mAP@IOU(\%)} \\
        \cmidrule(lr){2-7}
        & 0.1 & 0.2 & 0.3 & 0.4 & 0.5 & AVG \\
        \midrule
        RAD(2018) & 5.73 & 4.41 & 2.69 & 1.93 & 1.44 & 3.24 \\
        AVVD(2022) & 10.27 & 7.01 & 6.25 & 3.42 & 3.29 & 6.05 \\
        VadCLIP(2024) & 11.72 & 7.83 & 6.40 & 4.53 & 2.93 & 6.68\\
        ITC(2024) & 13.54 & 9.24 & 7.45 & 5.46 & 3.79 & 7.90 \\
        ReFLIP(2024) & 14.23 & 10.34 & 9.32 & 7.54 & 6.81 & 9.62\\
        % \rowcolor{gray!20}
        \textbf{DSANet(Ours)} & \textbf{21.39} & \textbf{14.96} & \textbf{11.74} & \textbf{8.98} & \textbf{8.00} & \textbf{13.01} \\
        \bottomrule
    \end{tabular}
    % \vspace{-2mm}
    \caption{Fine-grained comparisons on UCF-Crime.}
    \label{tab:fg_UCF}
\end{table}

\begin{table}[tb]
    \centering
    \small
    
    \setlength{\tabcolsep}{10.2pt}
    \begin{tabular}{ccc|cc}
        \toprule
        Adapter & SG-NM & DCSA & AP(\%) & AVG(\%) \\
        \midrule
        
        \multicolumn{3}{l|}{Baseline} & 84.51 & 24.70 \\
        \checkmark & & & 85.00 & 28.15 \\
        \checkmark & \checkmark & & 85.94 & 28.39 \\
        \checkmark & & \checkmark & 85.67 & 28.25 \\
        % \rowcolor{gray!20}
        \checkmark & \checkmark & \checkmark & \textbf{86.95} & \textbf{28.87} \\
        \bottomrule
    \end{tabular}
    % \vspace{-2mm}
    \caption{Ablation studies on model components. ``SG-NM'' denotes Self-Guided Normality Modeling, and ``DCSA'' denotes Decoupled Contrastive Semantic Alignment.}
    \label{tab:ablation_study1}
\end{table}

\begin{table}[tb]
    \centering
    \small
    
    \setlength{\tabcolsep}{20.0pt}
    \begin{tabular}{c|cc}
        \toprule
        Method & AP(\%) & AVG(\%) \\
        \midrule
        
        No tuning & 81.57 & 27.60 \\
        Manual Prompt & 81.05 & 28.05 \\
        Learning Prompt & 82.88 & 28.26 \\
        % \rowcolor{gray!20}
        \textbf{Ours(Adapter)} & \textbf{86.95} & \textbf{28.87} \\
        \bottomrule
    \end{tabular}
    % \vspace{-2mm}
    \caption{Effectiveness of Text Encoder Tuning.}
    \label{tab:ablation_study2}
\end{table}

\subsection{Main Results}

\noindent\textbf{Coarse-grained WS-VAD Results.}
Tables~\ref{tab:cg_XD} and~\ref{tab:cg_UCF} show coarse-grained results on XD-Violence and UCF-Crime. On XD-Violence, DSANet achieves \textbf{86.95\%} AP, surpassing ReFLIP (85.81\%) and ITC (85.45\%). On UCF-Crime, it reaches \textbf{89.44\%} AUC, outperforming ITC (89.04\%) and ReFLIP (88.57\%) with the same backbone. These results validate the strong reasoning capabilities of DSANet in coarse-level anomaly detection under weak supervision.

\noindent\textbf{Fine-grained WS-VAD Results.}
We assess DSANet’s classification and localization ability in fine-grained anomaly detection. As shown in Table~\ref{tab:fg_XD}, it consistently outperforms prior methods on XD-Violence across all IoUs, with an average mAP of \textbf{28.87\%}, surpassing ReFLIP (27.36\%) and ITC (26.83\%), demonstrating robustness in capturing temporal anomaly boundaries.
On the more challenging UCF-Crime (Table~\ref{tab:fg_UCF}), DSANet achieves the best performance with an average mAP of \textbf{13.01\%}, outperforming ReFLIP (9.62\%) and ITC (7.90\%), with clear gains across all IoU levels (\eg, 21.39\% at 0.1 and 8.00\% at 0.5).
These results confirm DSANet’s effectiveness in capturing cross-modal fine-grained semantics for precise anomaly classification and localization.

\subsection{Ablation Studies}
We conduct ablation studies on XD-Violence to dissect our model and validate the contribution of its key components.

\noindent\textbf{Effectiveness of Components.}
Table~\ref{tab:ablation_study1} shows module-wise impact. Based on the VadCLIP~\cite{wu2024vadclip} baseline, adding the Adapter improves AP / AVG to 84.75\% / 28.31\%, showing the value of task-specific text adaptation.
Introducing the Self-Guided Normality Modeling branch raises performance to 85.94\% / 28.84\%, showing its effectiveness in anomaly detection. Adding the Decoupled Contrastive Semantic Alignment mechanism yields 85.70\% / 28.60\%, confirming its role in semantic alignment for better localization and classification.
DSANet with all modules reaches 86.95\% / 28.87\%, surpassing baseline by 2.44\% AP and 4.17\% AVG, with strong component synergy.

\noindent\textbf{Effectiveness of Text Encoder Tuning.}
Table~\ref{tab:ablation_study2} compares CLIP text encoder adaptation strategies. Using a frozen encoder with class labels performs moderately (81.38\% AP, 27.60\% AVG). Manual prompt(``a video of \textless label\textgreater'') slightly improves AVG but reduces AP.
Learning prompt (\eg, CoOp-style) improves to (82.69\% AP, 28.66\% AVG), while our Adapter-based tuning performs best (86.95\% AP, 28.87\% AVG), confirming that internal adaptation offers more expressive and task-aligned features than others.

\subsection{Qualitative Results}
\begin{figure}[tb]
    
    \centering
    
    \includegraphics[width=1.0\columnwidth]{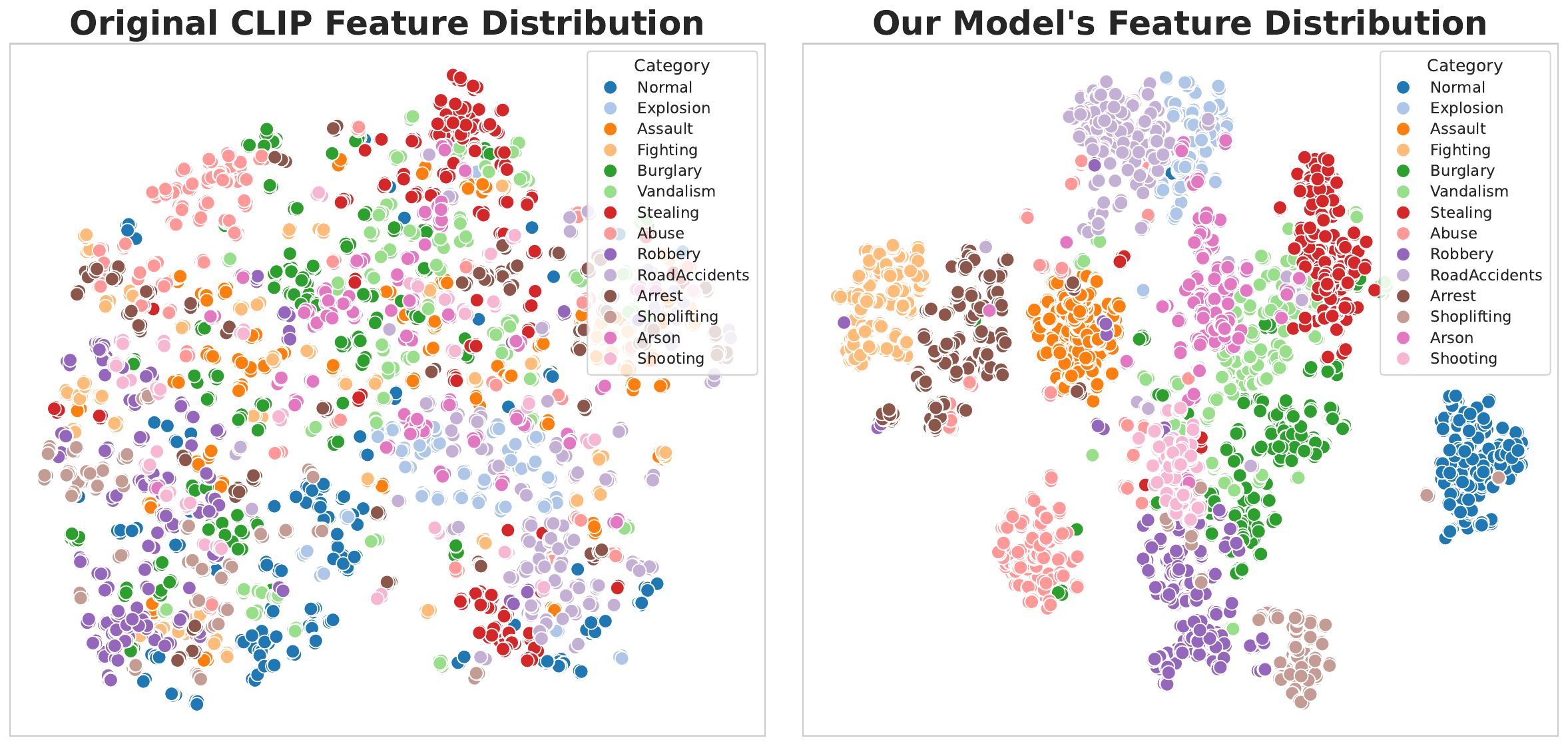}
    
    \caption{t-SNE visualizations for UCF-Crime.}
    \label{fig:tsne_ucf}
\end{figure}

\noindent\textbf{t-SNE Visualization of Category Separability.}
To assess feature discriminativeness, we visualize UCF-Crime using t-SNE in Figure~\ref{fig:tsne_ucf}. Compared to original CLIP features (left), which show a heavy category overlap, our model’s features (right) exhibit clearer inter-class boundaries and tighter intra-class clusters. This improved separability demonstrates that our model enhances class-discriminative representations, especially for visually similar anomaly categories.

\begin{figure}[tb]
    
    \centering
    
    \includegraphics[width=1.0\columnwidth]{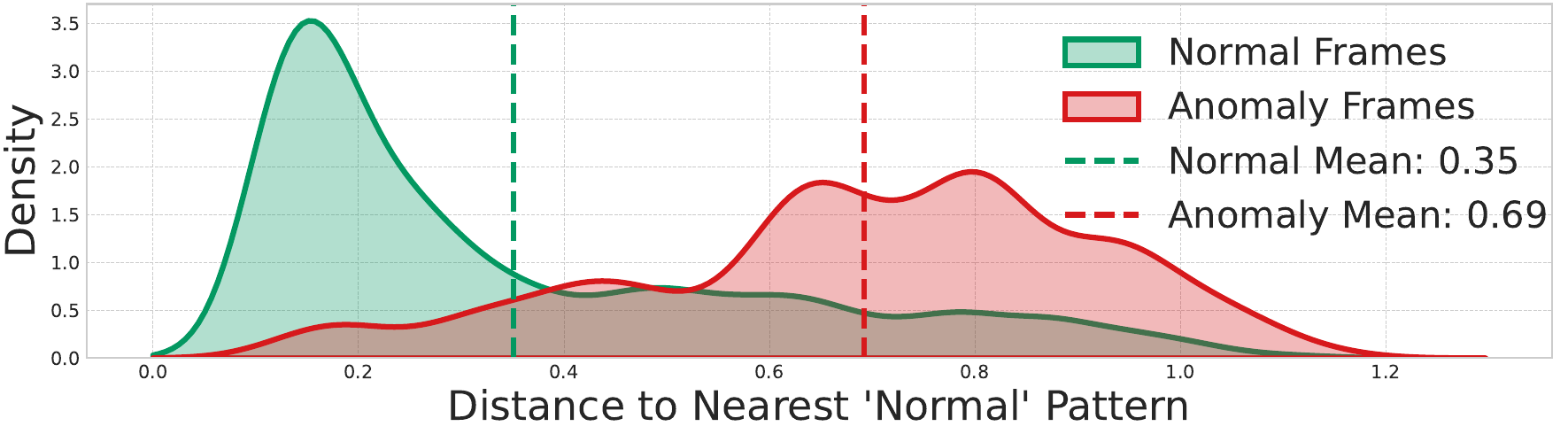}
    
    % \vspace{-2mm}
    % \caption{Comparison of Frame DNPs.}
    \caption{Comparison of Frame Distances to DNPs.}
    \label{fig:dnps}
\end{figure}
\begin{figure}[tb]
    
    \centering
    
    \includegraphics[width=1.0\columnwidth]{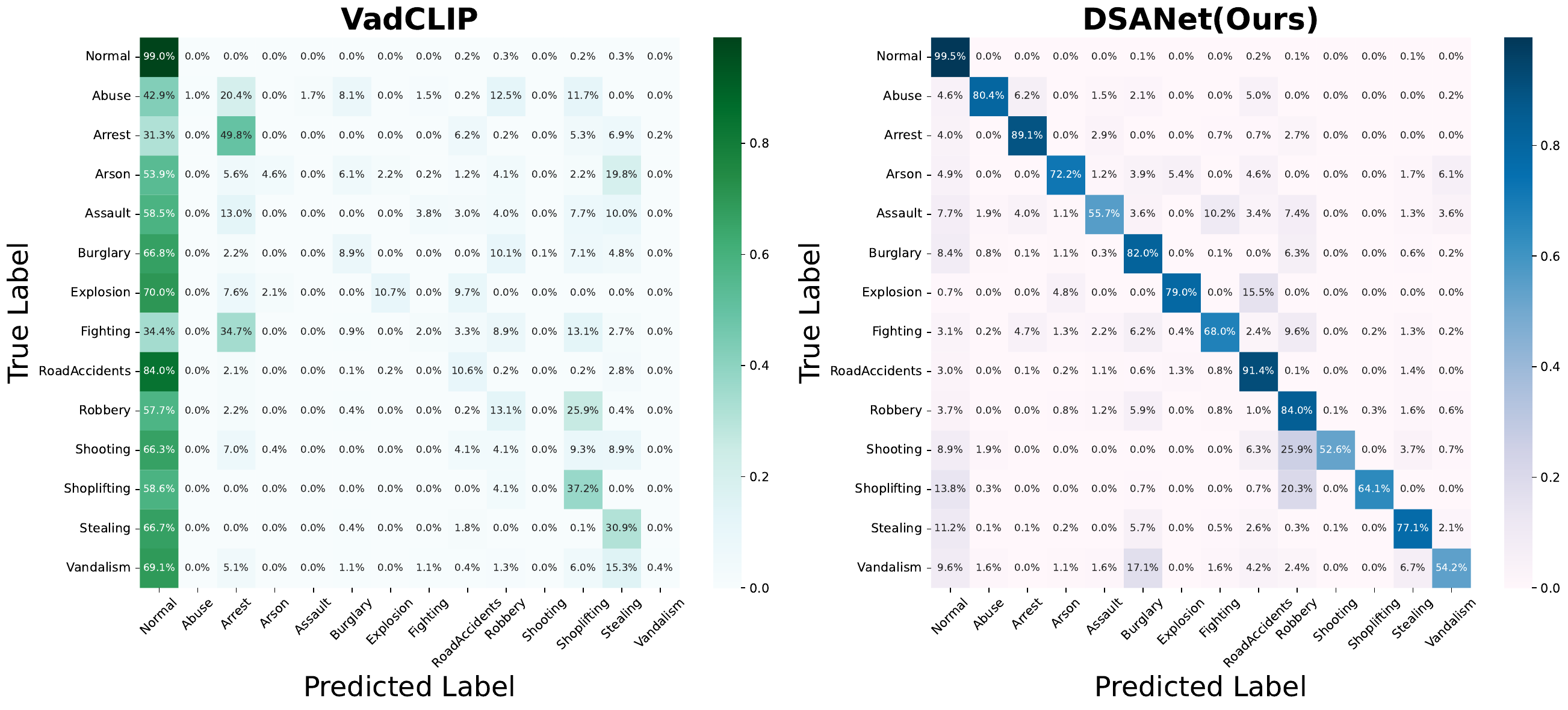}
    
    % \vspace{-2mm}
    \caption{
    Comparative results of semantic alignment matrix.
    % Comparison of Semantic Alignment Results.
    }
    \label{fig:event-centric}
\end{figure}
\begin{figure}[tb]
    
    \centering
    
    \includegraphics[width=1.0\columnwidth]{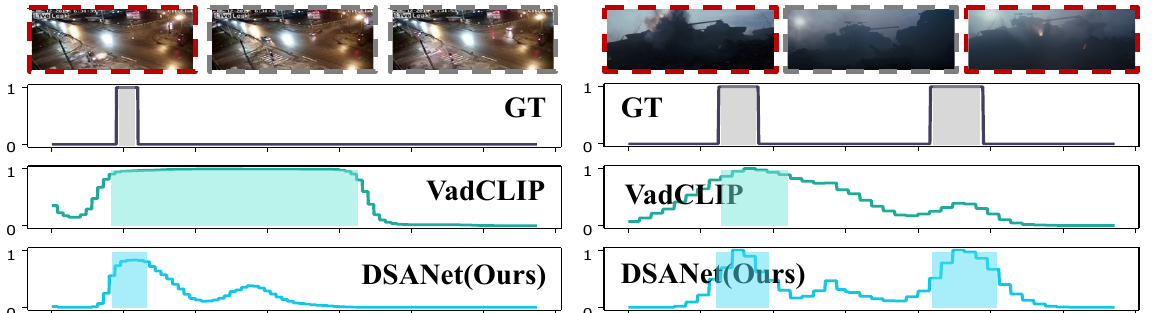}
    
    % \vspace{-2mm}
    \caption{Comparison of coarse-grained anomaly detection.}
    \label{fig:cg_with_vadclip}
\end{figure}

\noindent\textbf{Effects of Dynamic Normal Patterns.}
To evaluate the quality of our Dynamic Normal Patterns (DNPs), we measure the minimum cosine distance from each frame to its video's $K$ DNPs, assuming normal frames are closer to this DNP-defined space. We compute distributions of these distances across all frames in the test set.
Figure~\ref{fig:dnps} shows that normal frames concentrate at lower distances (mean: 0.35), while abnormal ones lie farther away (mean: 0.69), confirming that DNPs form a compact and discriminative representation of normal patterns.
This validates the reconstruction pathway guided by DNPs (Sec.~\ref{sec:SGNM}), where using only DNPs as key and value ensures high reconstruction errors for anomalous frames, making anomalies readily detectable.

\noindent\textbf{Effects of Decoupled Contrastive Semantic Alignment.}
We evaluate the Decoupled Contrastive Semantic Alignment(DCSA) by analyzing the alignment behavior of learned event-centric and background-centric prototypes. On UCF-Crime, we compare DSANet with VadCLIP~\cite{wu2024vadclip}, which is adapted to compute both prototypes using our formulation (Eq.~\ref{eq:dcsa}). For each video, its two prototypes are aligned with 14 class-level text embeddings, and the closest match is taken as the predicted label.
Figure~\ref{fig:event-centric} compares the confusion matrices of the event-centric prototype. VadCLIP shows severe misclassification among visually similar categories and often defaults to the ``normal'' class. In contrast, DSANet produces a much stronger diagonal dominance, reflecting improved class separability. For the background-centric prototype, DSANet achieves a 99.63\% alignment accuracy with ``normal'' class, outperforming VadCLIP’s 87.63\%, confirming its ability to isolate background information from event content.
These results demonstrate that DCSA effectively reduces class confusion, leading to more precise and disentangled representations for fine-grained anomaly detection.

\noindent\textbf{Results on Coarse-Grained Anomaly Detection.}
We visualize the anomaly detection results of DSANet and VadCLIP in Figure~\ref{fig:cg_with_vadclip}. As shown, VadCLIP often captures only the most salient anomaly segments, resulting in fragmented predictions and misaligned temporal boundaries. This aligns with our observation that MIL-only methods tend to focus on peak activation regions, leading to imprecise localization.
In contrast, DSANet produces predictions that align more accurately with the ground truth, accurately covering abnormal events while maintaining clear separation from normal regions. 
This improvement stems from leveraging learned normality prototypes, which guide the model to better capture temporal structure, thereby complementing the detection branch and enhancing localization precision.
These results confirm that explicitly modeling normal patterns is beneficial for stabilizing anomaly predictions and reducing boundary ambiguity in coarse-grained detection tasks.

% ----------- Conclusion -----------
\section{Conclusion}
In this paper, we present a novel framework named DSANet for WS-VAD that disentangles normal and anomaly semantics at both coarse-grained and fine-grained levels. By integrating self-guided normality modeling and decoupled contrastive semantic alignment, DSANet achieves improved temporal localization and class discrimination. Extensive experiments validate the efficacy of the proposed DSANet, achieving state-of-the-art  anomaly detection performance.

% ----------- Acknowledgement -----------
\section*{Acknowledgements}
This work is supported by the National Natural Science Foundation of China under grants  U22B2053 and 623B2039, and in part by the Interdisciplinary Research Program of HUST (2024JCYJ034).

\bibliography{aaai2026}

@article{ho2025review,
  title={A Review on Vision-Language-Based Approaches: Challenges and Applications.},
  author={Ho, Huu-Tuong and Nguyen, Luong Vuong and Pham, Minh-Tien and Pham, Quang-Huy and Tran, Quang-Duong and Huy, Duong Nguyen Minh and Nguyen, Tri-Hai},
  journal={Computers, Materials \& Continua},
  volume={82},
  number={2},
  year={2025}
}

@inproceedings{radford2021learning,
  title={Learning transferable visual models from natural language supervision},
  author={Radford, Alec and Kim, Jong Wook and Hallacy, Chris and Ramesh, Aditya and Goh, Gabriel and Agarwal, Sandhini and Sastry, Girish and Askell, Amanda and Mishkin, Pamela and Clark, Jack and others},
  booktitle={International conference on machine learning},
  pages={8748--8763},
  year={2021},
  organization={PmLR}
}

@inproceedings{jia2021scaling,
  title={Scaling up visual and vision-language representation learning with noisy text supervision},
  author={Jia, Chao and Yang, Yinfei and Xia, Ye and Chen, Yi-Ting and Parekh, Zarana and Pham, Hieu and Le, Quoc and Sung, Yun-Hsuan and Li, Zhen and Duerig, Tom},
  booktitle={International conference on machine learning},
  pages={4904--4916},
  year={2021},
  organization={PMLR}
}

@inproceedings{li2022blip,
  title={Blip: Bootstrapping language-image pre-training for unified vision-language understanding and generation},
  author={Li, Junnan and Li, Dongxu and Xiong, Caiming and Hoi, Steven},
  booktitle={International conference on machine learning},
  pages={12888--12900},
  year={2022},
  organization={PMLR}
}

@misc{yu2022cocacontrastivecaptionersimagetext,
      title={CoCa: Contrastive Captioners are Image-Text Foundation Models}, 
      author={Jiahui Yu and Zirui Wang and Vijay Vasudevan and Legg Yeung and Mojtaba Seyedhosseini and Yonghui Wu},
      year={2022},
      eprint={2205.01917},
      archivePrefix={arXiv},
      primaryClass={cs.CV},
      url={https://arxiv.org/abs/2205.01917}, 
}

@article{alayrac2022flamingo,
  title={Flamingo: a visual language model for few-shot learning},
  author={Alayrac, Jean-Baptiste and Donahue, Jeff and Luc, Pauline and Miech, Antoine and Barr, Iain and Hasson, Yana and Lenc, Karel and Mensch, Arthur and Millican, Katherine and Reynolds, Malcolm and others},
  journal={Advances in neural information processing systems},
  volume={35},
  pages={23716--23736},
  year={2022}
}

@inproceedings{li2023blip,
  title={Blip-2: Bootstrapping language-image pre-training with frozen image encoders and large language models},
  author={Li, Junnan and Li, Dongxu and Savarese, Silvio and Hoi, Steven},
  booktitle={International conference on machine learning},
  pages={19730--19742},
  year={2023},
  organization={PMLR}
}

@inproceedings{wang2024text,
  title={Text is mass: Modeling as stochastic embedding for text-video retrieval},
  author={Wang, Jiamian and Sun, Guohao and Wang, Pichao and Liu, Dongfang and Dianat, Sohail and Rabbani, Majid and Rao, Raghuveer and Tao, Zhiqiang},
  booktitle={Proceedings of the IEEE/CVF Conference on Computer Vision and Pattern Recognition},
  pages={16551--16560},
  year={2024}
}

@inproceedings{tian2024holistic,
  title={Holistic features are almost sufficient for text-to-video retrieval},
  author={Tian, Kaibin and Zhao, Ruixiang and Xin, Zijie and Lan, Bangxiang and Li, Xirong},
  booktitle={Proceedings of the IEEE/CVF Conference on Computer Vision and Pattern Recognition},
  pages={17138--17147},
  year={2024}
}

@inproceedings{zou2024language,
  title={Language-aware visual semantic distillation for video question answering},
  author={Zou, Bo and Yang, Chao and Qiao, Yu and Quan, Chengbin and Zhao, Youjian},
  booktitle={Proceedings of the IEEE/CVF Conference on Computer Vision and Pattern Recognition},
  pages={27113--27123},
  year={2024}
}

@inproceedings{li2024configure,
  title={How to configure good in-context sequence for visual question answering},
  author={Li, Li and Peng, Jiawei and Chen, Huiyi and Gao, Chongyang and Yang, Xu},
  booktitle={Proceedings of the IEEE/CVF Conference on Computer Vision and Pattern Recognition},
  pages={26710--26720},
  year={2024}
}

@misc{huang2024frosterfrozenclipstrong,
      title={FROSTER: Frozen CLIP Is A Strong Teacher for Open-Vocabulary Action Recognition}, 
      author={Xiaohu Huang and Hao Zhou and Kun Yao and Kai Han},
      year={2024},
      eprint={2402.03241},
      archivePrefix={arXiv},
      primaryClass={cs.CV},
      url={https://arxiv.org/abs/2402.03241}, 
}

@misc{jia2023generatingactionconditionedpromptsopenvocabulary,
      title={Generating Action-conditioned Prompts for Open-vocabulary Video Action Recognition}, 
      author={Chengyou Jia and Minnan Luo and Xiaojun Chang and Zhuohang Dang and Mingfei Han and Mengmeng Wang and Guang Dai and Sizhe Dang and Jingdong Wang},
      year={2023},
      eprint={2312.02226},
      archivePrefix={arXiv},
      primaryClass={cs.CV},
      url={https://arxiv.org/abs/2312.02226}, 
}

@inproceedings{sultani2018real,
  title={Real-world anomaly detection in surveillance videos},
  author={Sultani, Waqas and Chen, Chen and Shah, Mubarak},
  booktitle={Proceedings of the IEEE conference on computer vision and pattern recognition},
  pages={6479--6488},
  year={2018}
}

@inproceedings{tian2021weakly,
  title={Weakly-supervised video anomaly detection with robust temporal feature magnitude learning},
  author={Tian, Yu and Pang, Guansong and Chen, Yuanhong and Singh, Rajvinder and Verjans, Johan W and Carneiro, Gustavo},
  booktitle={Proceedings of the IEEE/CVF international conference on computer vision},
  pages={4975--4986},
  year={2021}
}

@inproceedings{lv2023unbiased,
  title={Unbiased multiple instance learning for weakly supervised video anomaly detection},
  author={Lv, Hui and Yue, Zhongqi and Sun, Qianru and Luo, Bin and Cui, Zhen and Zhang, Hanwang},
  booktitle={Proceedings of the IEEE/CVF conference on computer vision and pattern recognition},
  pages={8022--8031},
  year={2023}
}

@inproceedings{feng2021mist,
  title={Mist: Multiple instance self-training framework for video anomaly detection},
  author={Feng, Jia-Chang and Hong, Fa-Ting and Zheng, Wei-Shi},
  booktitle={Proceedings of the IEEE/CVF conference on computer vision and pattern recognition},
  pages={14009--14018},
  year={2021}
}

@inproceedings{joo2023clip,
  title={Clip-tsa: Clip-assisted temporal self-attention for weakly-supervised video anomaly detection},
  author={Joo, Hyekang Kevin and Vo, Khoa and Yamazaki, Kashu and Le, Ngan},
  booktitle={2023 IEEE International Conference on Image Processing (ICIP)},
  pages={3230--3234},
  year={2023},
  organization={IEEE}
}

@inproceedings{wu2024vadclip,
  title={Vadclip: Adapting vision-language models for weakly supervised video anomaly detection},
  author={Wu, Peng and Zhou, Xuerong and Pang, Guansong and Zhou, Lingru and Yan, Qingsen and Wang, Peng and Zhang, Yanning},
  booktitle={Proceedings of the AAAI Conference on Artificial Intelligence},
  volume={38},
  pages={6074--6082},
  year={2024}
}

@article{pu2024learning,
  title={Learning prompt-enhanced context features for weakly-supervised video anomaly detection},
  author={Pu, Yujiang and Wu, Xiaoyu and Yang, Lulu and Wang, Shengjin},
  journal={IEEE Transactions on Image Processing},
  year={2024},
  publisher={IEEE}
}

@article{liu2024injecting,
  title={Injecting text clues for improving anomalous event detection from weakly labeled videos},
  author={Liu, Tianshan and Lam, Kin-Man and Bao, Bing-Kun},
  journal={IEEE Transactions on Image Processing},
  year={2024},
  publisher={IEEE}
}

@inproceedings{zanella2024harnessing,
  title={Harnessing large language models for training-free video anomaly detection},
  author={Zanella, Luca and Menapace, Willi and Mancini, Massimiliano and Wang, Yiming and Ricci, Elisa},
  booktitle={Proceedings of the IEEE/CVF Conference on Computer Vision and Pattern Recognition},
  pages={18527--18536},
  year={2024}
}

@inproceedings{zhang2025holmes,
  title={Holmes-vau: Towards long-term video anomaly understanding at any granularity},
  author={Zhang, Huaxin and Xu, Xiaohao and Wang, Xiang and Zuo, Jialong and Huang, Xiaonan and Gao, Changxin and Zhang, Shanjun and Yu, Li and Sang, Nong},
  booktitle={Proceedings of the Computer Vision and Pattern Recognition Conference},
  pages={13843--13853},
  year={2025}
}

@inproceedings{wu2020not,
  title={Not only look, but also listen: Learning multimodal violence detection under weak supervision},
  author={Wu, Peng and Liu, Jing and Shi, Yujia and Sun, Yujia and Shao, Fangtao and Wu, Zhaoyang and Yang, Zhiwei},
  booktitle={European conference on computer vision},
  pages={322--339},
  year={2020},
  organization={Springer}
}

@article{wu2024deep,
  title={Deep learning for video anomaly detection: A review},
  author={Wu, Peng and Pan, Chengyu and Yan, Yuting and Pang, Guansong and Wang, Peng and Zhang, Yanning},
  journal={arXiv preprint arXiv:2409.05383},
  year={2024}
}

@article{abdalla2024video,
  title={Video anomaly detection in 10 years: A survey and outlook},
  author={Abdalla, Moshira and Javed, Sajid and Radi, Muaz Al and Ulhaq, Anwaar and Werghi, Naoufel},
  journal={arXiv preprint arXiv:2405.19387},
  year={2024}
}

@article{nayak2021comprehensive,
  title={A comprehensive review on deep learning-based methods for video anomaly detection},
  author={Nayak, Rashmiranjan and Pati, Umesh Chandra and Das, Santos Kumar},
  journal={Image and Vision Computing},
  volume={106},
  pages={104078},
  year={2021},
  publisher={Elsevier}
}

@inproceedings{chen2024prompt,
  title={Prompt-enhanced multiple instance learning for weakly supervised video anomaly detection},
  author={Chen, Junxi and Li, Liang and Su, Li and Zha, Zheng-jun and Huang, Qingming},
  booktitle={Proceedings of the IEEE/CVF Conference on Computer Vision and Pattern Recognition},
  pages={18319--18329},
  year={2024}
}

@inproceedings{carreira2017quo,
  title={Quo vadis, action recognition? a new model and the kinetics dataset},
  author={Carreira, Joao and Zisserman, Andrew},
  booktitle={proceedings of the IEEE Conference on Computer Vision and Pattern Recognition},
  pages={6299--6308},
  year={2017}
}

@inproceedings{luo2025exploring,
  title={Exploring intrinsic normal prototypes within a single image for universal anomaly detection},
  author={Luo, Wei and Cao, Yunkang and Yao, Haiming and Zhang, Xiaotian and Lou, Jianan and Cheng, Yuqi and Shen, Weiming and Yu, Wenyong},
  booktitle={Proceedings of the Computer Vision and Pattern Recognition Conference},
  pages={9974--9983},
  year={2025}
}

@inproceedings{hasan2016learning,
  title={Learning temporal regularity in video sequences},
  author={Hasan, Mahmudul and Choi, Jonghyun and Neumann, Jan and Roy-Chowdhury, Amit K and Davis, Larry S},
  booktitle={Proceedings of the IEEE conference on computer vision and pattern recognition},
  pages={733--742},
  year={2016}
}

@inproceedings{li2022self,
  title={Self-training multi-sequence learning with transformer for weakly supervised video anomaly detection},
  author={Li, Shuo and Liu, Fang and Jiao, Licheng},
  booktitle={Proceedings of the AAAI Conference on Artificial Intelligence},
  pages={1395--1403},
  year={2022}
}

@inproceedings{pu2022locality,
  title={Locality-aware attention network with discriminative dynamics learning for weakly supervised anomaly detection},
  author={Pu, Yujiang and Wu, Xiaoyu},
  booktitle={2022 IEEE International Conference on Multimedia and Expo (ICME)},
  pages={1--6},
  year={2022},
  organization={IEEE}
}

@inproceedings{zhou2023dual,
  title={Dual memory units with uncertainty regulation for weakly supervised video anomaly detection},
  author={Zhou, Hang and Yu, Junqing and Yang, Wei},
  booktitle={Proceedings of the AAAI Conference on Artificial Intelligence},
  pages={3769--3777},
  year={2023}
}

@inproceedings{yang2024text,
  title={Text prompt with normality guidance for weakly supervised video anomaly detection},
  author={Yang, Zhiwei and Liu, Jing and Wu, Peng},
  booktitle={Proceedings of the IEEE/CVF conference on computer vision and pattern recognition},
  pages={18899--18908},
  year={2024}
}

@article{dev2024reflip,
  title={Reflip-vad: Towards weakly supervised video anomaly detection via vision-language model},
  author={Dev, Prabhu Prasad and Hazari, Raju and Das, Pranesh},
  journal={IEEE Transactions on Circuits and Systems for Video Technology},
  year={2024},
  publisher={IEEE}
}

@InProceedings{Wang_2021_ICCV,
    author    = {Wang, Xiang and Zhang, Shiwei and Qing, Zhiwu and Shao, Yuanjie and Zuo, Zhengrong and Gao, Changxin and Sang, Nong},
    title     = {OadTR: Online Action Detection With Transformers},
    booktitle = {Proceedings of the IEEE/CVF International Conference on Computer Vision (ICCV)},
    month     = {October},
    year      = {2021},
    pages     = {7565-7575}
}

@InProceedings{Wang_2022_CVPR,
    author    = {Wang, Xiang and Zhang, Shiwei and Qing, Zhiwu and Tang, Mingqian and Zuo, Zhengrong and Gao, Changxin and Jin, Rong and Sang, Nong},
    title     = {Hybrid Relation Guided Set Matching for Few-Shot Action Recognition},
    booktitle = {Proceedings of the IEEE/CVF Conference on Computer Vision and Pattern Recognition (CVPR)},
    month     = {June},
    year      = {2022},
    pages     = {19948-19957}
}

@InProceedings{Yu_2025_CVPR,
    author    = {Yu, Chunlin and Wang, Hanqing and Shi, Ye and Luo, Haoyang and Yang, Sibei and Yu, Jingyi and Wang, Jingya},
    title     = {SeqAfford: Sequential 3D Affordance Reasoning via Multimodal Large Language Model},
    booktitle = {Proceedings of the IEEE/CVF Conference on Computer Vision and Pattern Recognition (CVPR)},
    month     = {June},
    year      = {2025},
    pages     = {1691-1701}
}

@article{wang2025affordance,
  title={Affordance-R1: Reinforcement Learning for Generalizable Affordance Reasoning in Multimodal Large Language Model},
  author={Wang, Hanqing and Wang, Shaoyang and Zhong, Yiming and Yang, Zemin and Wang, Jiamin and Cui, Zhiqing and Yuan, Jiahao and Han, Yifan and Liu, Mingyu and Ma, Yuexin},
  journal={arXiv preprint arXiv:2508.06206},
  year={2025}
}

@article{wang2025dag,
  title={DAG: Unleash the Potential of Diffusion Model for Open-Vocabulary 3D Affordance Grounding},
  author={Wang, Hanqing and Zhang, Zhenhao and Ji, Kaiyang and Liu, Mingyu and Yin, Wenti and Chen, Yuchao and Liu, Zhirui and Zeng, Xiangyu and Gui, Tianxiang and Zhang, Hangxing},
  journal={arXiv preprint arXiv:2508.01651},
  year={2025}
}

@article{wang2025sdeval,
  title={SDEval: Safety Dynamic Evaluation for Multimodal Large Language Models},
  author={Wang, Hanqing and Tian, Yuan and Liu, Mingyu and Zhang, Zhenhao and Zhu, Xiangyang},
  journal={arXiv preprint arXiv:2508.06142},
  year={2025}
}

@article{shi2025shield,
  title={Shield: An evaluation benchmark for face spoofing and forgery detection with multimodal large language models},
  author={Shi, Yichen and Gao, Yuhao and Lai, Yingxin and Wang, Hongyang and Feng, Jun and He, Lei and Wan, Jun and Chen, Changsheng and Yu, Zitong and Cao, Xiaochun},
  journal={Visual Intelligence},
  volume={3},
  number={1},
  pages={9},
  year={2025},
  publisher={Springer}
}

@article{zhu2024video,
  title={Video anomaly detection with long-and-short-term time series correlations},
  author={Zhu, Xinrui and Qian, Xiaoyan and Shi, Yuzhou and Tao, Xudong and Li, Zhiyu},
  journal={Journal of Image and Graphics},
  volume={29},
  number={7},
  pages={1998--2010},
  year={2024},
  doi={10.11834/jig.230406}
}

@article{liang2023video,
  title={Video anomaly detection by fusing self-attention and autoencoder},
  author={Liang, Jiafei and Li, Ting and Yang, Jiaqi and Li, Yanan and Fang, Zhiwen and Yang, Feng},
  journal={Journal of Image and Graphics},
  volume={28},
  number={4},
  pages={1029--1040},
  year={2023},
  doi={10.11834/jig.211147}
}

\end{document}